\documentclass{article}

\usepackage{arxiv}

\usepackage[utf8]{inputenc}
\usepackage[T1]{fontenc}
\usepackage{hyperref}
\usepackage{url}
\usepackage{booktabs}
\usepackage{amsfonts}
\usepackage{nicefrac}
\usepackage[expansion=false]{microtype}
\usepackage{graphicx}
\usepackage{float}
\usepackage[section]{placeins}
\usepackage{amsmath,amssymb,amsthm}
\usepackage{multirow}
\usepackage{array}
\usepackage[dvipsnames]{xcolor}
\usepackage{caption}
\usepackage{enumitem}
\usepackage{subcaption}
\usepackage[capitalize]{cleveref}
\usepackage{setspace}
\usepackage{natbib}
\usepackage{doi}
\usepackage{adjustbox}
\usepackage{pifont}
\usepackage{tabularx}
\usepackage{mathrsfs}
\usepackage{bm}
\usepackage{colortbl}
\usepackage{textcomp}
\usepackage{threeparttable}

\usepackage{siunitx}
\title{CucumberVision at the Edge: Accuracy, Throughput, and Energy of RGB-D Cucumber Sizing on Jetson Orin Nano and Raspberry Pi 4}

\author{
  Manveen Kaur \\
  Department of Mechanical Engineering\\
  University of Windsor\\
  Windsor, ON, Canada \\
  \texttt{Kaur79@uwindsor.ca}
  \And
  Rajmeet Singh$^{*}$ \\
  Mohamed bin Zayed University of Artificial Intelligence\\
  Abu Dhabi, UAE \\
  \texttt{rajmeet.bhourji@mbzuai.ac.ae}
  \And
  Saeed Mozaffari \\
  Department of Mechanical Engineering\\
  University of Windsor\\
  Windsor, ON, Canada
  \And
  Majid Ahmadi \\
  Department of Electrical and Computer Engineering\\
  University of Windsor\\
  Windsor, ON, Canada
  \And
  Shahpour Alirezaee \\
  Department of Mechanical Engineering\\
  University of Windsor\\
  Windsor, ON, Canada
  \And
  \vspace{-1em}
  \normalsize\textit{$^{*}$ Corresponding author.}
}

\hypersetup{
  pdftitle={CucumberVision at the Edge: Accuracy, Throughput, and Energy of RGB-D Cucumber Sizing on Jetson Orin Nano and Raspberry Pi 4},
  pdfsubject={Edge Computing, Embedded Vision, Precision Agriculture},
  pdfauthor={Manveen Kaur, Rajmeet Singh, Saeed Mozaffari, Majid Ahmadi, Shahpour Alirezaee},
  pdfkeywords={Edge computing, embedded vision, post-training quantization, TensorRT, energy per inference, precision agriculture, RGB-D sensing},
  colorlinks=true,
  citecolor=blue,
  linkcolor=blue,
  urlcolor=blue
}

\begin{document}
\maketitle

\begin{abstract}
Non-contact fruit sizing from RGB-D imagery reaches sub-centimetre accuracy on workstation hardware, but in-field grading requires the same pipeline to run on a low-power single-board computer, and the cost of that transition is rarely reported. We deploy a curvature-aware cucumber length-measurement pipeline YOLO26n detection, prompt-based segmentation and a medial arc spline estimator on an NVIDIA Jetson Orin Nano and a Raspberry Pi~4, and characterise both against an RTX~3050~Ti workstation by replaying 50 identical greenhouse captures of 7 specimens, with specimen-level cluster-bootstrap statistics. Accuracy is preserved: mean absolute error (MAE) is 0.58~cm (95\% CI 0.46--0.69) on both the workstation and the Jetson at matched precision, with a paired difference of zero. TensorRT FP16 compilation cuts Jetson detection latency $2.57\times$, giving two operating points on one board: 72~ms per fruit at 0.84~cm, or 2.74~s at 0.64~cm. Four ablations map the design space. For prompted segmentation backends, measurement accuracy is fixed by the backend and reproduces across platforms to within 0.01~cm, while its cost is fixed by the platform and varies $214\times$, so the two choices are separable. Whole-graph INT8 quantization of the NMS-free detection head fails structurally, returning no detections, because one output tensor carries both pixel coordinates and confidences; we show that no calibration strategy can avoid this. Excluding the head restores function at $1.37\times$. Energy per fruit is 42.2~J on the Jetson, and its 7~W mode consumes 73\% more than its 15~W mode. The Raspberry Pi with MobileSAM costs $2.5\times$ more per fruit than the Jetson with SAM despite drawing 47\% less power, and it runs without thermal throttling under sustained load. The results define an accuracy--throughput--energy envelope with three non-dominated configurations and give rules for selecting among them.
\end{abstract}

\keywords{Edge computing \and embedded vision \and post-training quantization \and TensorRT \and energy per inference \and precision agriculture \and RGB-D sensing}

\section{Introduction}

Fruit length is the primary grading criterion in commercial greenhouse cucumber production, and non-contact measurement from RGB-D imagery offers a route to automating it. Depth-camera sizing has been demonstrated for several crops~\citep{genemola2021, genemola2023}, and our previous originating study established that a curvature-aware estimator fitting a cubic spline through the three-dimensional cross-section centroid spine of a segmented fruit and integrating arc length attains sub-centimetre accuracy under canopy conditions~\citep{kaur2026}. That result was obtained on workstation hardware with a discrete GPU.

Practical grading does not take place at a workstation. A scouting platform moving between rows carries a single-board computer with a few watts of power budget, no discrete GPU and, in the cheapest case, no active cooling. Whether a method that works on a laptop also works there is not a foregone conclusion: prompt-based segmentation may be one to two orders of magnitude more expensive on embedded silicon~\citep{kirillov2023, zhang2023}, reduced-precision arithmetic may or may not preserve the geometry the estimator depends on~\citep{wu2020}, and sustained inference on a passively cooled board may throttle. Existing edge benchmarks in agriculture report detection accuracy and frame rate on a single platform~\citep{alqahtani2024, sapkota2025}, but not the joint accuracy--throughput--energy behaviour of a complete measurement pipeline, and not the interaction between segmentation backend and platform.

This paper reports the deployment of that pipeline on two representative embedded platforms an NVIDIA Jetson Orin Nano and a Raspberry Pi~4 Model~B and characterises the resulting design space through four controlled ablations. All three platforms replay the same greenhouse captures against the same ground truth, so every reported difference is attributable to the deployment configuration rather than to sensing. The contributions are:
\begin{enumerate}
    \item Working deployments of the complete measurement pipeline on a Jetson Orin Nano and a Raspberry Pi~4, with accuracy shown to be statistically indistinguishable from the workstation baseline, together with the TensorRT and ONNX Runtime inference paths that make them practical.
    \item A segmentation-backend ablation establishing that, for prompted backends, measurement accuracy is determined by the mask source and transfers between platforms, while segmentation cost is determined by the platform and does not making backend selection and hardware selection separable design decisions.
    \item Identification of a structural, rather than numerical, failure mode in post-training INT8 quantization of NMS-free detection heads, with a proof that no calibration strategy can avoid it, a fix, and its measured cost.
    \item A characterisation of the deployment design space spline smoothing, power mode, energy per measurement and sustained thermal behaviour yielding an explicit accuracy--throughput--energy envelope and rules for selecting an operating point within it.
\end{enumerate}

\section{Related Work}

\subsection{Edge Deployment of Deep Learning in Precision Agriculture}

Deploying convolutional detectors on single-board computers is now routine in agricultural vision, and comparative benchmarks exist. \citet{alqahtani2024} evaluate YOLOv8, EfficientDet-Lite and SSD families across Raspberry Pi~3, 4 and~5, with and without Coral TPU accelerators, and across Jetson Nano and Jetson Orin Nano, reporting inference time, accuracy and energy. \citet{sapkota2025} benchmark the YOLO26 family against earlier YOLO versions and transformer detectors on Jetson platforms, with explicit attention to edge deployment readiness.

These studies characterise the detector in isolation. An applied measurement pipeline additionally carries a segmentation stage and a geometric estimator, and on the hardware considered here the detector is not always the dominant cost. Reporting detector latency alone therefore does not predict end-to-end throughput, and detection accuracy alone does not predict measurement accuracy.

\subsection{Post-Training Quantization for Detection Models}

Post-training quantization maps floating-point tensors to low-precision integers using a scale and zero-point recovered from calibration data, and is attractive at the edge because it needs neither retraining nor labelled data~\citep{jacob2018, nagel2021}. The governing design choice is granularity. \citet{wu2020} analyse per-tensor against per-channel scaling and show that per-tensor quantization degrades sharply when a tensor's dynamic range is heterogeneous, because a single scale must serve every element; their recommended workflow keeps accuracy within 1\% of the floating-point baseline across a wide range of models.

That literature treats the failure as a matter of degree accuracy falls by some measurable amount and addresses it with better calibration or finer granularity. NMS-free architectures such as RT-DETR~\citep{zhao2024} and YOLOv10~\citep{wang2024yolov10} concatenate coordinates and scores into a single output tensor, a design choice whose interaction with per-tensor quantization has not been analysed. The failure reported in \cref{sec:quantization} is different in kind: the quantized model returns no detections at all, and we show the cause is a range conflict internal to that tensor that no calibration strategy and no weight-granularity change can resolve.

\subsection{Foundation-Model Segmentation on Embedded Hardware}

The Segment Anything Model (SAM)~\citep{kirillov2023} produces high-quality masks from box or point prompts without task-specific training, which suits a pipeline whose detector already supplies boxes. SAM~2~\citep{ravi2024} extends the framework to video with a streaming memory architecture, though its edge deployment characteristics remain unreported. The encoder is expensive, and a family of distilled alternatives has followed: MobileSAM~\citep{zhang2023} replaces the ViT-H encoder with a lightweight one and reports a roughly sixty-fold parameter reduction, and EfficientSAM~\citep{xiong2024} uses masked image pretraining to recover accuracy at reduced cost.

These models are characterised by parameter count and by throughput on server or mobile GPUs. What has not been reported is how the choice interacts with the deployment platform in an applied pipeline: whether a distilled model on weak hardware is preferable to a full model on stronger hardware, and whether the accuracy a backend delivers on a workstation is the accuracy it delivers in the field. \Cref{sec:segmentation} addresses both.

\subsection{Non-Contact Fruit Sizing and Grading}

RGB-D and photogrammetric sizing has been studied most thoroughly for apples. \citet{miranda2023} survey AI-based fruit sizing methods and identify the progression from hand-crafted features to deep multitask architectures, noting that most validated systems target approximately spherical fruit. \citet{genemola2021} compare four methods for in-field size estimation from photogrammetry-derived point clouds under occlusion, \citet{genemola2023} extend this to amodal segmentation so that partially hidden fruit can still be sized, and \citet{ferrerferrer2023} demonstrate simultaneous detection and size estimation with a single multitask network. These works target approximately spherical fruit, for which a diameter or a fitted sphere is an adequate model.

Elongated and curved fruit such as cucumber are not well served by that assumption: a straight-line extent systematically under-reads a curved specimen, and the error grows with curvature. The originating study for the present work addresses this with a medial arc spline estimator, summarised in \cref{sec:pipeline}. Its accuracy was established on workstation hardware; the present paper asks what survives deployment.

\subsection{The Gap}

No prior study has characterised the joint accuracy--throughput--energy envelope of a complete RGB-D measurement pipeline for elongated and curved fruit across embedded platforms, nor established whether segmentation-backend choice and platform choice can be made independently, nor identified the structural failure mode of post-training INT8 quantization in NMS-free detection heads. This paper addresses all three.

\section{Materials and Methods}

\subsection{Measurement Pipeline Overview}
\label{sec:pipeline}

The pipeline is unchanged from the originating study~\citep{kaur2026}. An Intel RealSense D435 provides aligned colour and depth. A YOLO26n instance-segmentation model locates each fruit; a segmentation backend refines the detection to a pixel-level mask; five length estimators then operate on the resulting masked point cloud: a scan-line baseline over the bounding box (M1), principal component analysis (PCA) on the bounding-box cloud (M2), PCA on the masked cloud (M3), a keypoint-guided hybrid (M4) and the proposed medial arc spline (M5). Deprojection uses colour-stream intrinsics after alignment.

\begin{figure}[!ht]
\centering
\includegraphics[width=0.95\linewidth]{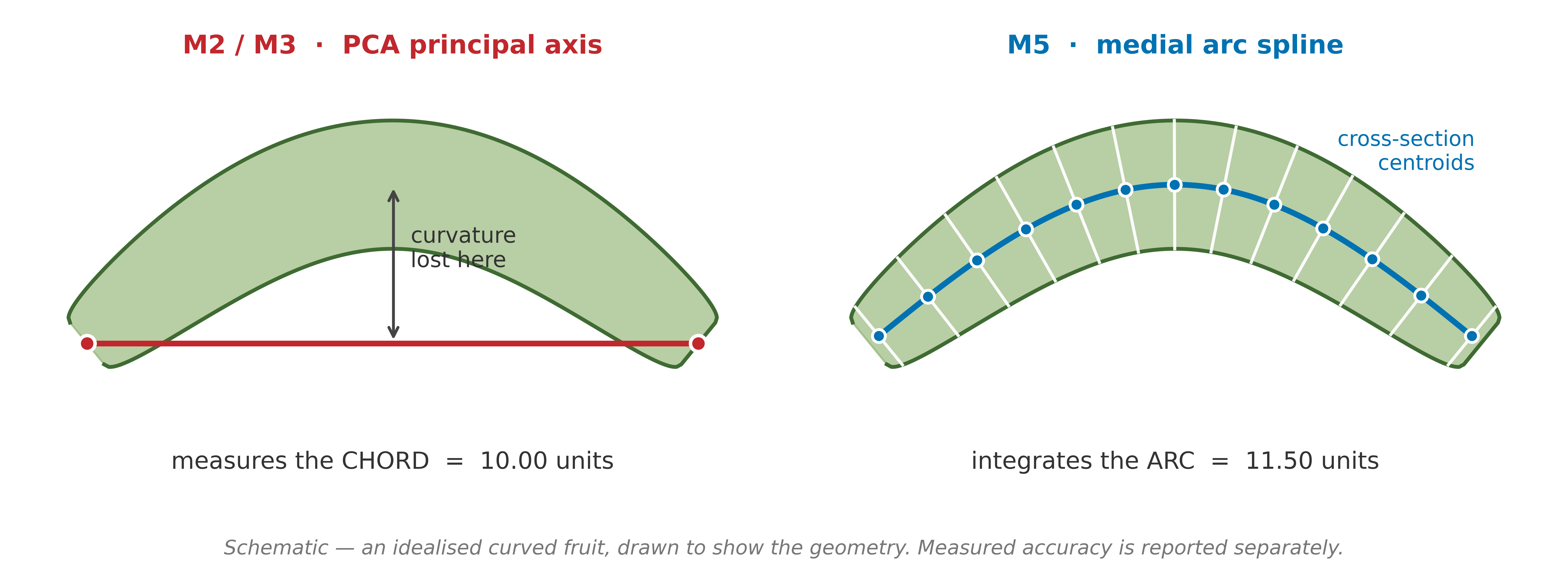}
\caption{Medial arc spline estimator (M5), schematic. Left: a straight principal axis measures the chord between the extremes and under-reads a curved fruit. Right: M5 fits a cubic spline through the cross-section centroid spine and integrates arc length. On the idealised shape drawn here the chord under-reads the arc by 13\%, and the discrepancy grows with curvature. }
\label{fig:fig1}
\end{figure}
One structural detail governs the interpretation of the ablations. M1 and M2 require only the detection bounding box and M4 only keypoints, whereas M3 and M5 require a pixel-level mask. M1 and M2 therefore act as controls on the detector: when a configuration change moves them, it changed the boxes; when it leaves them alone but moves M3 and M5, it acted through the segmentation stage instead. Results are reported for M5 unless stated otherwise (\cref{fig:fig1}).

\subsection{Hardware Platforms}

\begin{figure}[!ht]
\centering
\includegraphics[width=0.85\linewidth]{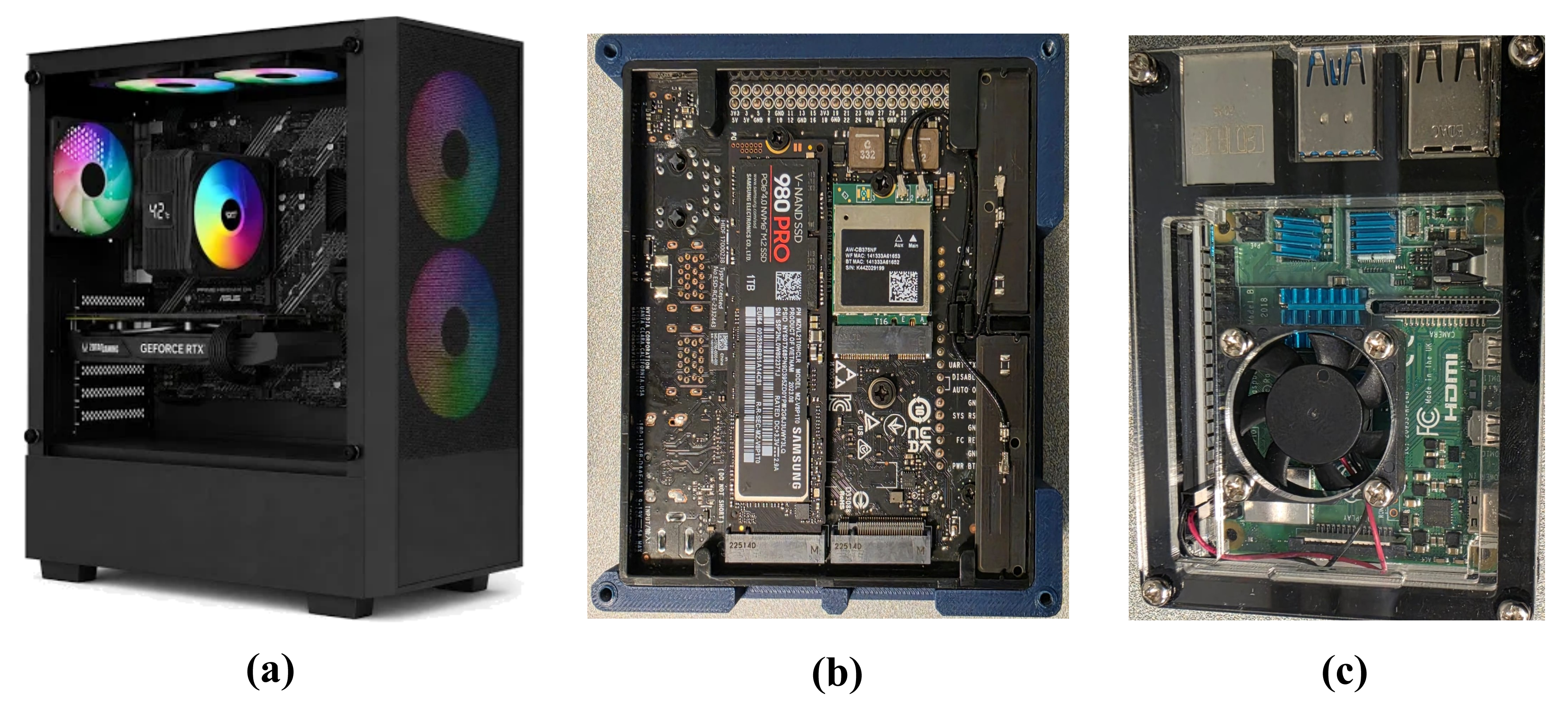}
\caption{Deployment platforms: (a) RTX 3050 Ti, (b) Jetson Orin Nano, and (c) Raspberry Pi 4 B.}
\label{fig:fig2}
\end{figure}

\cref{fig:fig2} and \Cref{tab:platforms} illustrated and summarises the three platforms used in this research work. The workstation, a laptop with an RTX~3050~Ti GPU (labelled ``Laptop'' in the result tables), serves as the accuracy baseline; the two embedded boards are the deployment targets (Jetson Orin Nano and Raspberry Pi 4 B).

\begin{table}[!ht]
\centering
\caption{Deployment platforms. All three replay identical stored captures.}
\label{tab:platforms}
\renewcommand{\arraystretch}{1.15}
\footnotesize
\begin{tabularx}{\linewidth}{@{}>{\raggedright\arraybackslash}p{1.2cm}*{3}{>{\raggedright\arraybackslash}X}@{}}
\toprule
 & \textbf{Workstation} & \textbf{Jetson Orin Nano} & \textbf{RPi~4 B} \\
\midrule
Compute & RTX 3050~Ti & Ampere, 1024 cores & Cortex-A72 CPU \\
Runtime & PyTorch CUDA & PyTorch / TensorRT & ONNX Runtime \\
Precision & FP32 & FP32, FP16 & FP32, INT8 \\
Software & --- & JetPack 6.0, TensorRT 8.6.2 & RPi OS 64-bit \\
Cooling & Active & Active & Passive \\
\bottomrule
\end{tabularx}
\end{table}

\subsection{Deployment and Inference Paths}

Except where marked ($\dagger$) in the result tables, both embedded deployments execute the detector through a direct inference path that loads the exported model and parses its output tensor explicitly via ONNX Runtime on the Raspberry Pi and the TensorRT Python API on the Jetson rather than through a general-purpose inference wrapper. YOLO26 is NMS-free~\citep{sapkota2025} and emits a tensor of shape $(1, 300, 38)$ that is already detections-first and pre-filtered, a layout the common wrappers do not assume. Parsing it directly guarantees correct interpretation and avoids post-processing the model does not require.

On the Jetson the detector is compiled from the same ONNX graph into a TensorRT FP16 engine. On the Raspberry Pi it is executed by ONNX Runtime in FP32 or in one of the quantized configurations of \cref{sec:quantization_design}.

\subsection{Benchmark Protocol}

\begin{figure}[!ht]
\centering
\includegraphics[width=0.85\linewidth]{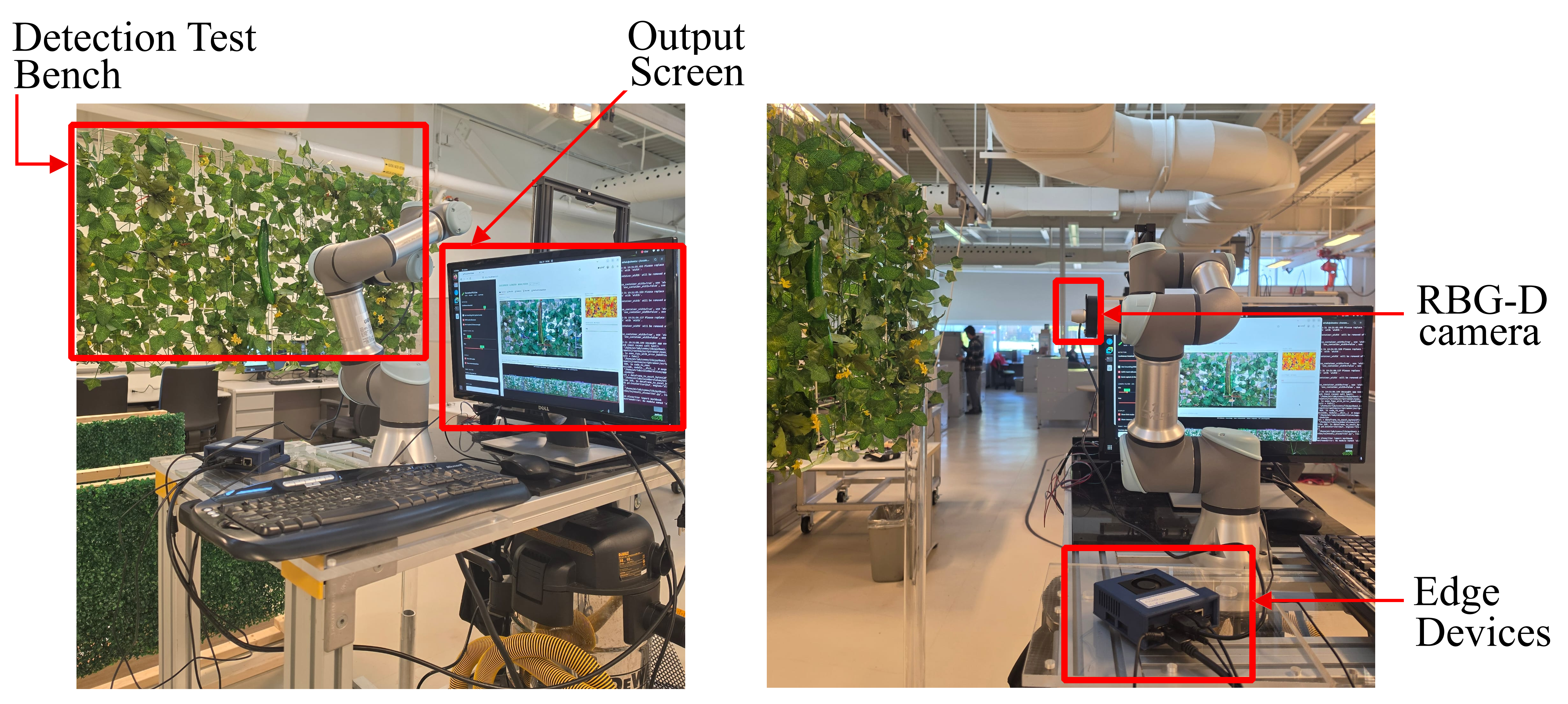}
\caption{Experimentation setup using Edge devices.}
\label{fig:fig2.1}
\end{figure}

All platforms replay the same 50 stored greenhouse captures from two sessions, comprising repeated views of physical specimens with paired thread ground-truth length spanning 6.9--25.0~cm. Also fresh In-lab experimentation are conducted using both edge devices, the RGB-D camera is mounted on the robot end and scan the cucumber canopy for real time length estimation (\cref{fig:fig2.1}). 

Latency is computed over 220 post-warm-up captures and accuracy over all 50, with both counts recorded in every summary the harness produces. Detection latency is the forward pass plus pre- and post-processing; segmentation latency is the mask-refinement pass; measurement latency is the estimator itself.

\subsection{Ablation Design}
\label{sec:quantization_design}

Four ablations are reported, each varying one factor with the others held fixed.

\textbf{Segmentation backend.} Three mask sources SAM ViT-B, the distilled MobileSAM, and the detector's own YOLO-seg masks were run on a single GPU with hardware constant, then each was re-run on the platform where it is practical, to test whether accuracy transfers. GrabCut was included on the Raspberry Pi as a classical reference point.

\textbf{Detector quantization.} The detector was exported to ONNX and quantized statically in QDQ format using calibration data from 40 held-out greenhouse frames. Four strategies were compared: per-tensor MinMax, per-tensor entropy, per-tensor percentile and per-channel MinMax.

\textbf{Spline smoothing.} The medial arc spline optionally applies a 1-2-1 smoothing pass to the centroid spine before fitting. The flag was toggled with hardware, detector and segmentation backend held fixed.

\textbf{Power mode.} The Jetson was benchmarked in its 15~W and 7~W modes with an otherwise identical TensorRT configuration.

\subsection{Energy and Thermal Instrumentation}

The Jetson carries INA3221 monitors exposed through \texttt{tegrastats}, giving board power without external hardware. Energy per measurement is obtained by integrating the VDD\_IN rail over the active window and dividing by the number of fruit measured. An idle baseline is recorded separately for each power mode and subtracted to give a net figure. The Raspberry Pi has no on-board power sensor; its supply power was measured externally, and energy and net energy were computed in the same way against a measured idle baseline of 2.52~W.

Sustained-load thermal behaviour was characterised on the Raspberry Pi by running consecutive benchmarks for 30~min while sampling SoC temperature, ARM clock and hardware throttle flags every 5~s.

\subsection{Statistical Analysis}

The 50 captures are repeated views of physical specimens, so captures are not independent observations. All interval estimates therefore use a cluster bootstrap~\citep{field2007} that resamples specimens rather than captures, with 10{,}000 replicates and percentile intervals. Paired configuration contrasts use an exact sign-flip permutation test in which all captures of a specimen change sign together, preserving the cluster structure. Agreement with ground truth is additionally summarized in Bland--Altman terms~\citep{bland1986} as bias and 95\% limits of agreement (LoA).

\section{Results}

\subsection{Deployment and Accuracy}

Both embedded platforms run the complete pipeline end to end. \Cref{tab:deployment} summarises latency, accuracy and memory across all configurations; \cref{tab:accuracy} gives the corresponding cluster-bootstrap confidence intervals and Bland--Altman agreement statistics.

On the workstation, SAM ViT-B yields the lowest M5 error at 0.58~cm (95\% CI 0.46--0.69), followed by MobileSAM at 0.76~cm (95\% CI 0.56--1.00) and native YOLO-seg masks at 0.98~cm (95\% CI 0.66--1.24). The Jetson Orin Nano at FP32 with SAM ViT-B reproduces the workstation result exactly (0.58~cm, bias $+0.06$, 95\% LoA $-1.68$ to $+1.81$). TensorRT FP16 compilation reduces detection latency from 120~ms to 46~ms while M5 increases marginally to 0.64~cm (95\% CI 0.55--0.73). With native masks the Jetson achieves the fastest per-fruit time of 0.07~s at 0.84~cm (95\% CI 0.69--0.98).

\begin{table}[!ht]
\centering
\caption{Cross-platform deployment results. Rows marked $\dagger$ used the library wrapper for detection.}
\label{tab:deployment}
\renewcommand{\arraystretch}{1.12}
\resizebox{\linewidth}{!}{%
\begin{tabular}{lrrrrrr}
\toprule
\textbf{Configuration} & \textbf{Det.} & \textbf{Seg.} & \textbf{Per fruit} & \textbf{MAE} & \textbf{RMSE} & \textbf{RAM} \\
 & \textbf{(ms)} & \textbf{(ms)} & \textbf{(s)} & \textbf{(cm)} & \textbf{(cm)} & \textbf{(MB)} \\
\midrule
Laptop, SAM ViT-B & 43 & 421 & 0.47 & 0.58 & 0.88 & 1466 \\
Laptop, MobileSAM & 46 & 70 & 0.12 & 0.76 & 1.13 & 1568 \\
Laptop, native & 46 & --- & 0.05 & 0.98 & 1.29 & 2059 \\
Jetson FP32, SAM & 120 & 2653 & 2.79 & 0.58 & 0.88 & 1272 \\
Jetson TRT FP16, SAM & 46 & 2680 & 2.74 & 0.64 & 0.97 & 1344 \\
Jetson TRT FP16, native & 55 & --- & 0.07 & 0.84 & 1.18 & 1027 \\
RPi FP32, MobileSAM$^\dagger$ & 3266 & 15041 & 18.34 & 0.76 & 1.10 & 1006 \\
RPi FP32, native & 1929 & --- & 1.96 & 0.82 & 1.18 & 682 \\
\bottomrule
\end{tabular}%
}
\end{table}

On the Raspberry Pi~4, MobileSAM attains 0.76~cm (95\% CI 0.55--0.94, $n = 49$) and native masks 0.82~cm (95\% CI 0.66--0.98), both with per-fruit latencies of 18.34~s and 1.96~s respectively. INT8 quantization with the detection head kept at FP32 degrades M5 to 1.12~cm (95\% CI 0.73--1.75) and widens the Bland--Altman limits of agreement to $-3.43$ to $+3.69$~cm. Configurations using native masks on both platforms show a positive Bland--Altman bias of approximately $+0.57$~cm, indicating a systematic tendency to overestimate length when the segmentation backend is bypassed.

\begin{table}[!ht]
\centering
\caption{Accuracy with cluster-bootstrap intervals and Bland--Altman agreement.}
\label{tab:accuracy}
\renewcommand{\arraystretch}{1.12}
\resizebox{\linewidth}{!}{%
\begin{tabular}{lcrcrr}
\toprule
\textbf{Configuration} & $n$ & \textbf{MAE (cm)} & \textbf{95\% CI} & \textbf{Bias} & \textbf{95\% LoA} \\
\midrule
Laptop, SAM ViT-B & 50 & 0.58 & 0.46--0.69 & +0.06 & $-1.68$ to $+1.81$ \\
Laptop, MobileSAM & 50 & 0.76 & 0.56--1.00 & $-0.02$ & $-2.26$ to $+2.22$ \\
Laptop, native & 50 & 0.98 & 0.66--1.24 & $-0.13$ & $-2.67$ to $+2.40$ \\
Jetson FP32, SAM & 50 & 0.58 & 0.46--0.69 & +0.06 & $-1.68$ to $+1.81$ \\
Jetson TRT FP16, SAM & 50 & 0.64 & 0.55--0.73 & +0.09 & $-1.83$ to $+2.01$ \\
Jetson TRT FP16, native & 50 & 0.84 & 0.69--0.98 & +0.57 & $-1.48$ to $+2.62$ \\
RPi FP32, native & 50 & 0.82 & 0.66--0.98 & +0.58 & $-1.45$ to $+2.61$ \\
RPi INT8 head-FP32 & 50 & 1.12 & 0.73--1.75 & +0.13 & $-3.43$ to $+3.69$ \\
RPi FP32, MobileSAM & 49 & 0.76 & 0.55--0.94 & $-0.49$ & $-2.44$ to $+1.46$ \\
\bottomrule
\end{tabular}%
}
\end{table}

\subsection{TensorRT Acceleration on the Jetson}

Compiling the same ONNX graph to a TensorRT FP16 engine reduces Jetson detection latency from 120~ms to 46~ms, a $2.57\times$ improvement, with segmentation and measurement unchanged. M5 moves from 0.58~cm to 0.64~cm. The paired difference is $+0.061$~cm (95\% CI $+0.005$ to $+0.137$, $p = 0.11$): consistently signed but not statistically significant at 7 specimens. The engine build warned that 115 weights fell outside the FP16 representable range; this proved benign, with outputs matching the FP32 path to within 1~px in box coordinates and 0.001 in confidence.

The accelerated detector makes two operating points available on one board: 72~ms per fruit at 0.84~cm using native masks, or 2.74~s at 0.64~cm using SAM. A $38\times$ latency difference buys 0.20~cm of accuracy.

\subsection{Ablation: Segmentation Backend}
\label{sec:segmentation}

Segmenter choice was isolated on one GPU with hardware held constant: SAM ViT-B gives 0.58~cm (95\% CI 0.46--0.69), MobileSAM 0.76~cm (95\% CI 0.56--1.00) and native YOLO-seg masks 0.98~cm (95\% CI 0.66--1.24). The ordering is monotonic in both cost and accuracy.

\begin{figure}[!ht]
\centering
\includegraphics[width=0.8\linewidth]{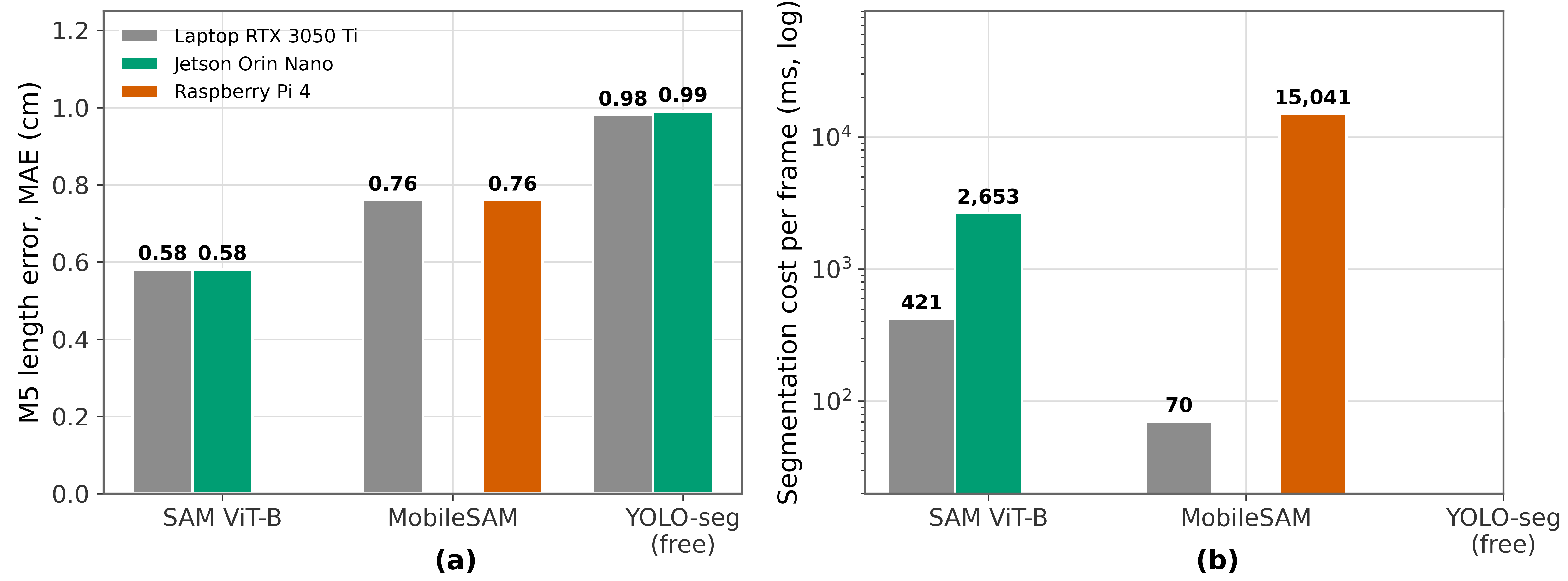}
\caption{Segmentation backend separability. (a)~M5 length error (MAE) is determined by the mask source and reproduces across platforms to within 0.01~cm. (b)~Segmentation cost per frame varies by up to $214\times$ across platforms for the same backend (log scale). YOLO-seg masks are free as they are produced during detection.}
\label{fig:fig4}
\end{figure}

The prompted backends reproduce across platforms. Comparing like with like FP32 on both sides the largest laptop-to-embedded disagreement for SAM and MobileSAM is 0.01~cm, and neither paired contrast approaches significance. This is the result that licenses choosing a prompted segmentation backend on a workstation and expecting the accuracy to hold in the field. Native masks do not reproduce in the same way: M5 with native masks is 0.98~cm on the laptop but 0.82~cm on the Raspberry Pi and 0.84~cm on the Jetson, consistent with the lower cross-platform mask agreement reported in \cref{sec:agreement}. \cref{fig:fig4} illustrated the segmentation backend separability for M5 length error and cost per frame. 

Cost does not transfer at all. MobileSAM runs in 70~ms on the laptop and 15{,}041~ms on the Raspberry Pi, a factor of 214. Backend selection and platform selection are therefore separable design decisions: the backend determines the accuracy achievable, the platform determines the throughput at which it is achievable.

The classical reference point sets a lower bound. GrabCut costs $2.4\times$ less than MobileSAM but fails to produce a usable mask on 22\% of frames and degrades M5 to 1.92~cm---worse than the bounding-box estimator M2 at 0.83~cm, which never required a mask.

\subsection{Ablation: Detector Quantization}
\label{sec:quantization}

On the Raspberry Pi, post-training quantization applied to the whole graph reduces model size from 41.7 to 11.1~MB, cuts peak memory from 682 to 460~MB and reduces detection latency from 1{,}929 to 793~ms, a $2.43\times$ speed-up the fastest detector measured on the Raspberry Pi in this study (\cref{tab:quantization}). It also returns no detections, on every capture, on two platforms and two ONNX Runtime versions.

The cause is structural. The NMS-free detection head emits a single tensor whose columns hold box coordinates spanning 0--640 alongside a confidence spanning 0--1. Per-tensor quantization assigns that tensor one scale, necessarily set by the coordinates: the resulting step is 7.71. A peak confidence of 0.85 is 0.11 of one step and rounds to zero, so no candidate clears the detection threshold.

No calibration strategy can avoid this. Representing a 0.85 confidence as non-zero requires a step below 1.63; representing a coordinate of 640 in 255 unsigned levels requires a step of at least 2.51. The two requirements are mutually exclusive.

\begin{figure}[!ht]
\centering
\includegraphics[width=0.8\linewidth]{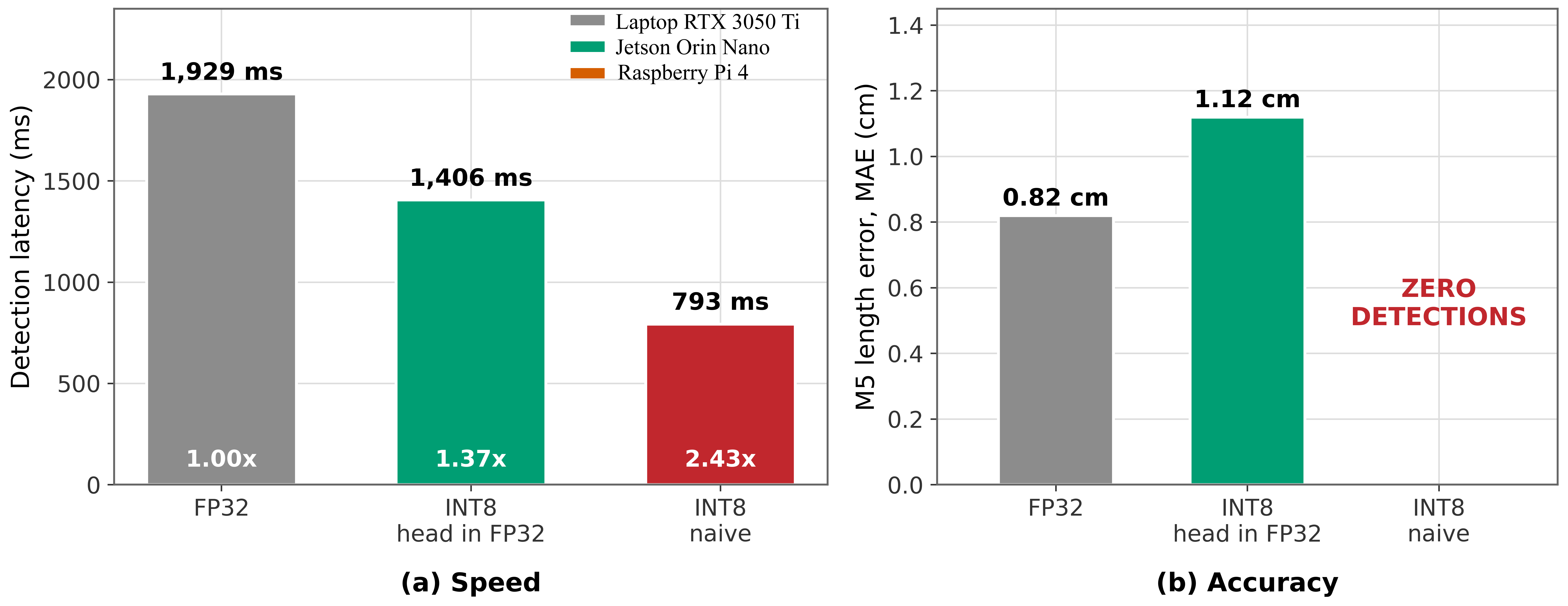}
\caption{Detector quantization on the Raspberry Pi 4. (a) Detection latency by configuration, annotated with speed-up over the FP32 direct path. (b) Resulting M5 error; the whole-graph INT8 configuration has no bar because it returns no detections on any capture despite being the fastest configuration in (a).}
\label{fig:fig5}
\end{figure}

Excluding the detection head from quantization resolves it. Detection latency becomes 1{,}406~ms, a $1.37\times$ speed-up, and M5 error moves from 0.82~cm to 1.12~cm ($+0.30$~cm, 95\% CI $-0.03$ to $+0.83$, $p = 0.27$) (\cref{fig:fig5}).

\begin{table}[!ht]
\centering
\caption{Detector quantization ablation on the Raspberry Pi~4.}
\label{tab:quantization}
\renewcommand{\arraystretch}{1.12}
\resizebox{\linewidth}{!}{%
\begin{tabular}{lrrlrrr}
\toprule
\textbf{Configuration} & \textbf{Det.} & \textbf{Speed-} & \textbf{Det-} & \textbf{MAE} & \textbf{Model} & \textbf{RAM} \\
 & \textbf{(ms)} & \textbf{up} & \textbf{ects} & \textbf{(cm)} & \textbf{(MB)} & \textbf{(MB)} \\
\midrule
FP32, wrapper & 3266 & $0.59\times$ & yes & --- & 41.7 & --- \\
FP32, direct & 1929 & $1.00\times$ & yes & 0.82 & 41.7 & 682 \\
INT8, head FP32 & 1406 & $1.37\times$ & yes & 1.12 & 15.1 & 594 \\
INT8, whole graph & 793 & $2.43\times$ & no & --- & 11.1 & 460 \\
\bottomrule
\end{tabular}%
}
\end{table}

\subsection{Ablation: Spline Smoothing}

The medial arc spline optionally smooths the centroid spine before fitting. On clean SAM masks smoothing costs $+0.07$~cm ($p = 0.44$), raising M5 from 0.58 to 0.65~cm, because it attenuates real curvature the mask had resolved correctly. On degenerate rectangle masks it recovers 3.10~cm. The practical rule is that smoothing should follow the segmentation backend rather than act as a global default---disabled with a learned segmenter, enabled where mask quality is uncertain or degraded.

\subsection{Ablation: Power Mode and Energy Per Measurement}
\label{sec:energy}

\Cref{tab:energy_edge} reports energy per measurement for each instrumented configuration. At its 15~W setting the Jetson draws 10.14~W while measuring and consumes 42.2~J per fruit gross, 20.2~J net of idle. A 100~Wh battery therefore supports approximately 8{,}500 measurements, which is not a limiting figure for a shift-length scouting run.

Repeating the identical configuration at 7~W gives the result that justifies measuring energy rather than power~\citep{yan2023}. The board draws 29\% less power but takes $2.44\times$ longer, so the work costs 73.1~J per fruit 73\% more than at 15~W. For this workload the lower power mode is the less energy-efficient one.

\begin{figure}[!ht]
\centering
\includegraphics[width=0.8\linewidth]{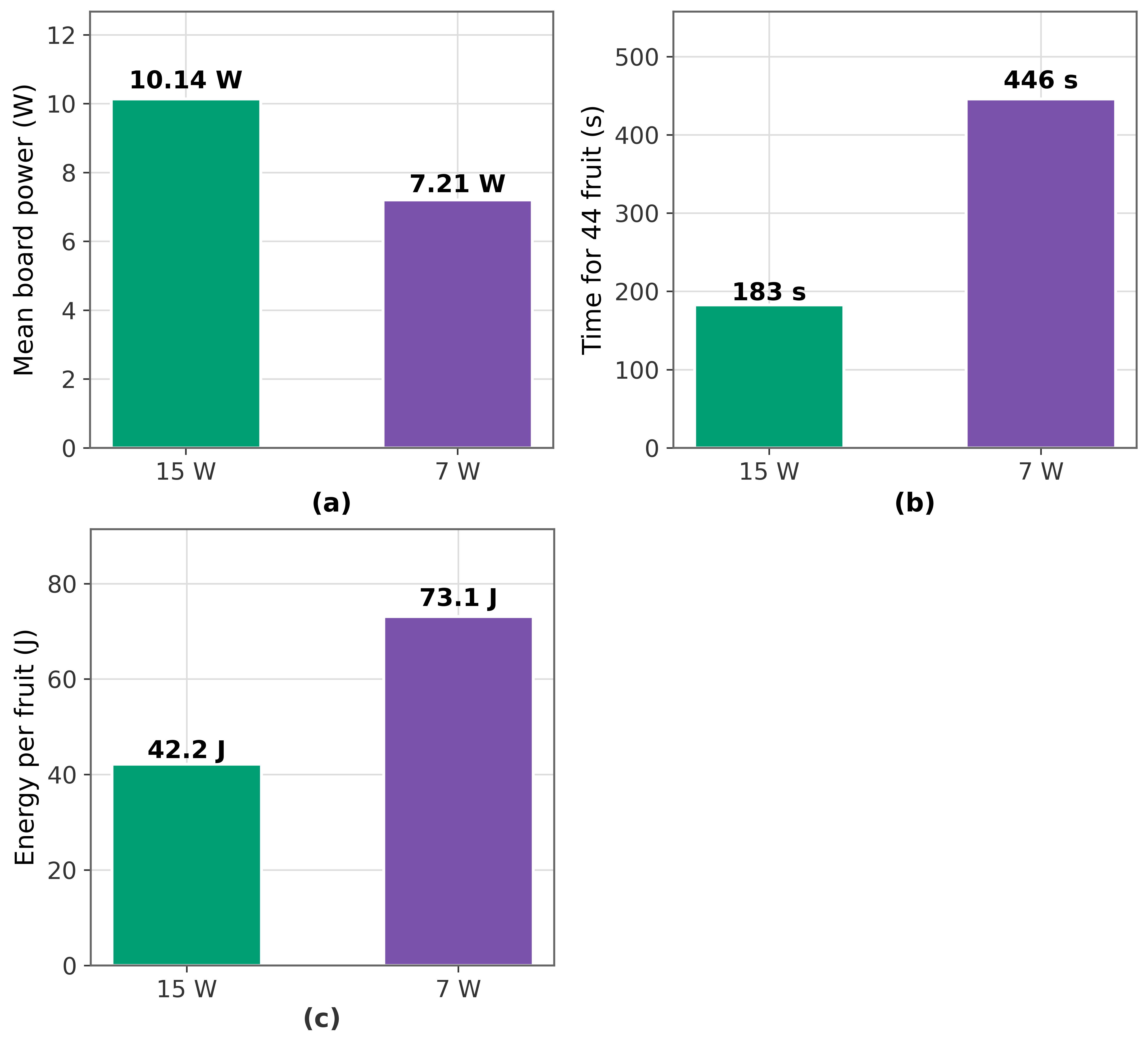}
\caption{Power-mode ablation on the Jetson Orin Nano, TensorRT FP16 with SAM. (a) Mean board draw, (b) time to measure 44 fruit, (c) resulting energy per fruit. Lower power draw does not imply lower energy per unit of work when the work takes proportionally longer.}
\label{fig:fig6}
\end{figure}

\Cref{fig:fig6} depicted the power mode ablation study on the Jetson Orin Nano. Accuracy is bit-identical in both modes (M5 0.64~cm in each; paired difference exactly zero), which serves as a control: a power cap changes execution speed, not arithmetic. The energy difference is entirely the time power trade, and it is concentrated in the segmentation stage, which slows by $2.75\times$ against $1.49\times$ for detection.

Measuring the same quantity on the Raspberry Pi turns the power-mode result into a cross-platform one. The Pi draws 5.41~W running MobileSAM against the Jetson's 10.14~W with SAM, 47\% less power, and costs 105.5~J per fruit against 42.2~J $2.50\times$ as much gross and $2.79\times$ net. These are the two configurations closest in accuracy (0.76~cm against 0.64~cm), and the lower-powered board is both slower and more expensive per measurement. The trade that makes the Jetson's 7~W mode wasteful operates between platforms as well as within one.

On the Pi, load power is nearly constant across configurations 5.41 to 5.68~W, a 5\% spread so energy there is execution time multiplied by a near-fixed draw, which makes any latency optimisation an energy optimisation of the same size. Excluding the detection head from quantization cuts detection latency $1.37\times$ and energy per fruit from 15.05 to 11.29~J, 25\% less. The Jetson does not behave this way, because its power mode changes the draw as well as the duration.

\begin{table}[!ht]
\centering
\caption{Energy per measurement across platforms, 44 fruit per run.}
\label{tab:energy_edge}
\small
\begin{threeparttable}
\begin{tabular}{@{}lccccc@{}}
\toprule
\textbf{Configuration} &
\textbf{\shortstack{Mean draw\\(W)}} &
\textbf{\shortstack{Run time\\(s)}} &
\textbf{J/fruit} &
\textbf{\shortstack{J/fruit,\\net}} &
\textbf{\shortstack{M5 MAE\\(cm)}} \\
\midrule
Jetson TensorRT FP16 + SAM, 15\,W   & 10.14 & 183 & 42.18  & 20.18 & 0.64 \\
Jetson TensorRT FP16 + SAM, 7\,W    & 7.21  & 446 & 73.12  & 23.45 & 0.64 \\
RPi~4 FP32 + MobileSAM              & 5.41  & 859 & 105.55 & 56.38 & 0.76 \\
RPi~4 FP32, native masks            & 5.68  & 117 & 15.05  & 8.38  & 0.82 \\
RPi~4 INT8 head-FP32, native masks  & 5.61  & 89  & 11.29  & 6.22  & 1.12 \\
\bottomrule
\end{tabular}
\begin{tablenotes}
\footnotesize
\item Raspberry Pi idle baseline: 2.52\,W; Jetson idle baselines are recorded separately for each power mode. Energy runs use a single timing repeat; the harness default of five multiplies the joules per fruit by the repeat count.
\end{tablenotes}
\end{threeparttable}
\end{table}

\subsection{Sustained Thermal Behaviour}
\label{sec:thermal}

Under 30~min of continuous inference the passively cooled Raspberry Pi reached thermal equilibrium within approximately 90~s and peaked at 53.0\,$^\circ$C, 27.0\,$^\circ$C below the 80\,$^\circ$C throttle threshold. The ARM clock held 1800~MHz across all 357 samples and no throttle flag was set. Per-run detection latency varied by 1.1\% non-monotonically, consistent with measurement noise rather than progressive degradation.

\begin{figure}[!ht]
\centering
\includegraphics[width=0.8\linewidth]{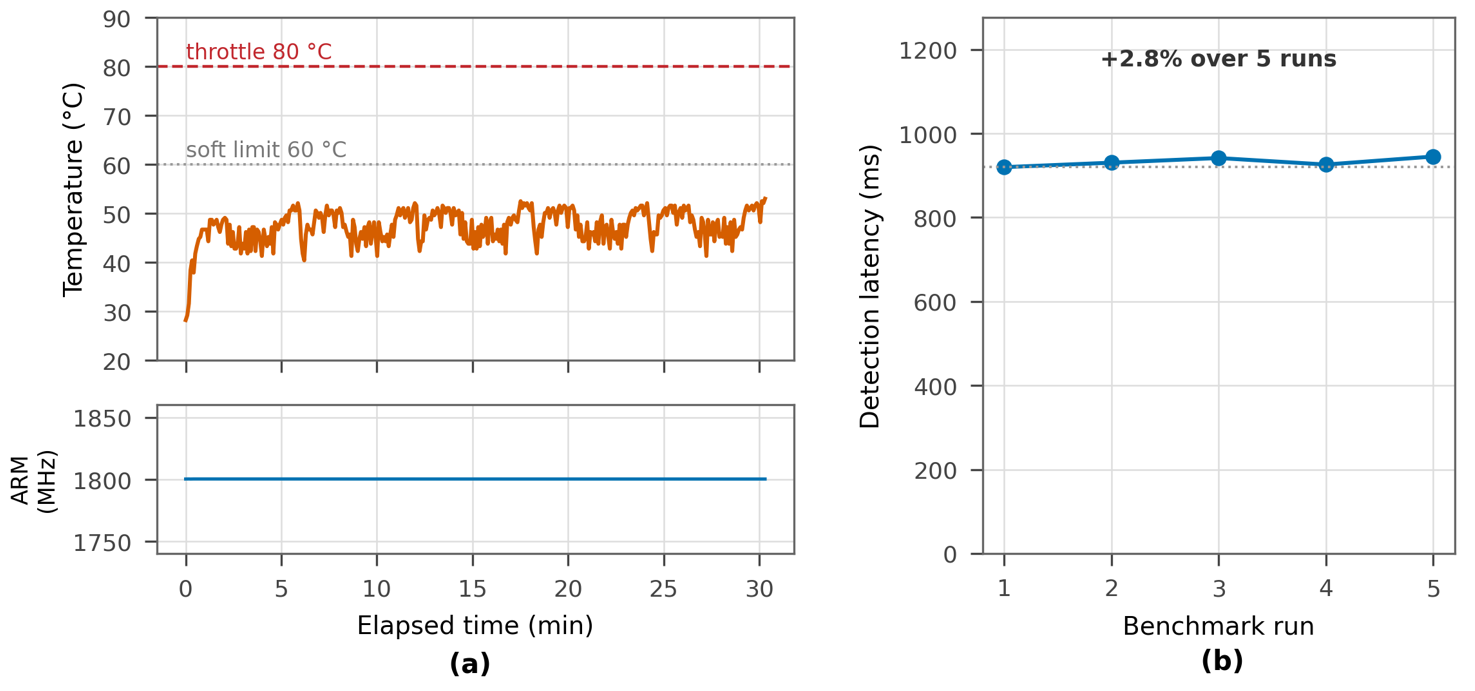}
\caption{Sustained-load thermal behaviour of the quantized pipeline on the Raspberry Pi 4 over 30 min of continuous inference at 21 °C ambient. (a) SoC temperature and ARM clock; the clock is held at its nominal 1800 MHz throughout. (b) Detection latency per run, varying non-monotonically by 1.1\%.}
\label{fig:fig7}
\end{figure}

\subsection{Detection and Mask Agreement Across Platforms}
\label{sec:agreement}

Every functional configuration (all except whole-graph INT8) detected the fruit on every capture: the detection rate is 100\% throughout. \Cref{tab:agreement} quantifies agreement against a reference run on five metrics.

Detection boxes are highly stable. The Jetson at FP32 reproduces the laptop's boxes exactly (IoU 1.000), and even after TensorRT FP16 compilation the worst-case box IoU remains 0.946, with a mean centroid shift of only 0.69~px. The Raspberry Pi shows comparable box stability (mean IoU 0.994, minimum 0.966, shift 0.54~px), confirming that the NMS-free head produces consistent detections regardless of platform or runtime.

Prompted segmentation masks are equally reproducible. SAM ViT-B yields bit-identical masks on the Jetson at FP32 (IoU 1.000), and MobileSAM on the Raspberry Pi matches the laptop to IoU 1.000 (minimum 0.997). TensorRT FP16 introduces minor rounding that lowers the minimum mask IoU to 0.992, but the mean remains 0.999. This near-perfect cross-platform mask agreement explains why M5 accuracy transfers between platforms to within 0.01~cm for prompted backends (\cref{sec:segmentation}).

Native YOLO-seg masks tell a different story. Cross-platform native mask IoU drops to a mean of 0.823 with a minimum of 0.447, meaning the worst-case mask disagrees with its reference on more than half its area. Even on the same laptop, native masks agree with SAM at only 0.789 mean IoU (minimum 0.398). The native segmentation head is therefore the dominant source of measurement variability in configurations that bypass prompted segmentation, accounting for the higher bias ($+0.57$~cm) observed in native-mask configurations in \cref{tab:accuracy}.

\begin{table}[!ht]
\centering
\caption{Detection and mask agreement against a reference run.}
\label{tab:agreement}
\renewcommand{\arraystretch}{1.12}
\resizebox{\linewidth}{!}{%
\begin{tabular}{lrrrr}
\toprule
\textbf{Comparison} & \textbf{Det.} & \textbf{Box IoU} & \textbf{Mask IoU} & \textbf{Shift} \\
 & \textbf{(\%)} & \textbf{mean (min)} & \textbf{mean (min)} & \textbf{px} \\
\midrule
Jetson FP32 vs laptop, SAM & 100 & 1.000 (1.000) & 1.000 (1.000) & 0.00 \\
Jetson TRT vs Jetson FP32 & 100 & 0.990 (0.946) & 0.999 (0.992) & 0.69 \\
RPi vs laptop, MobileSAM & 100 & 0.994 (0.966) & 1.000 (0.997) & 0.54 \\
RPi vs laptop, native & 100 & 0.994 (0.966) & 0.823 (0.447) & 0.54 \\
MobileSAM vs SAM, laptop & 100 & 1.000 (1.000) & 0.957 (0.788) & 0.00 \\
Native vs SAM, laptop & 100 & 1.000 (1.000) & 0.789 (0.398) & 0.00 \\
\bottomrule
\end{tabular}%
}
\end{table}

\section{Discussion}

\subsection{Configuration Selection and the Accuracy--Throughput--Energy Envelope}

\begin{figure}[!ht]
\centering
\includegraphics[width=0.95\linewidth]{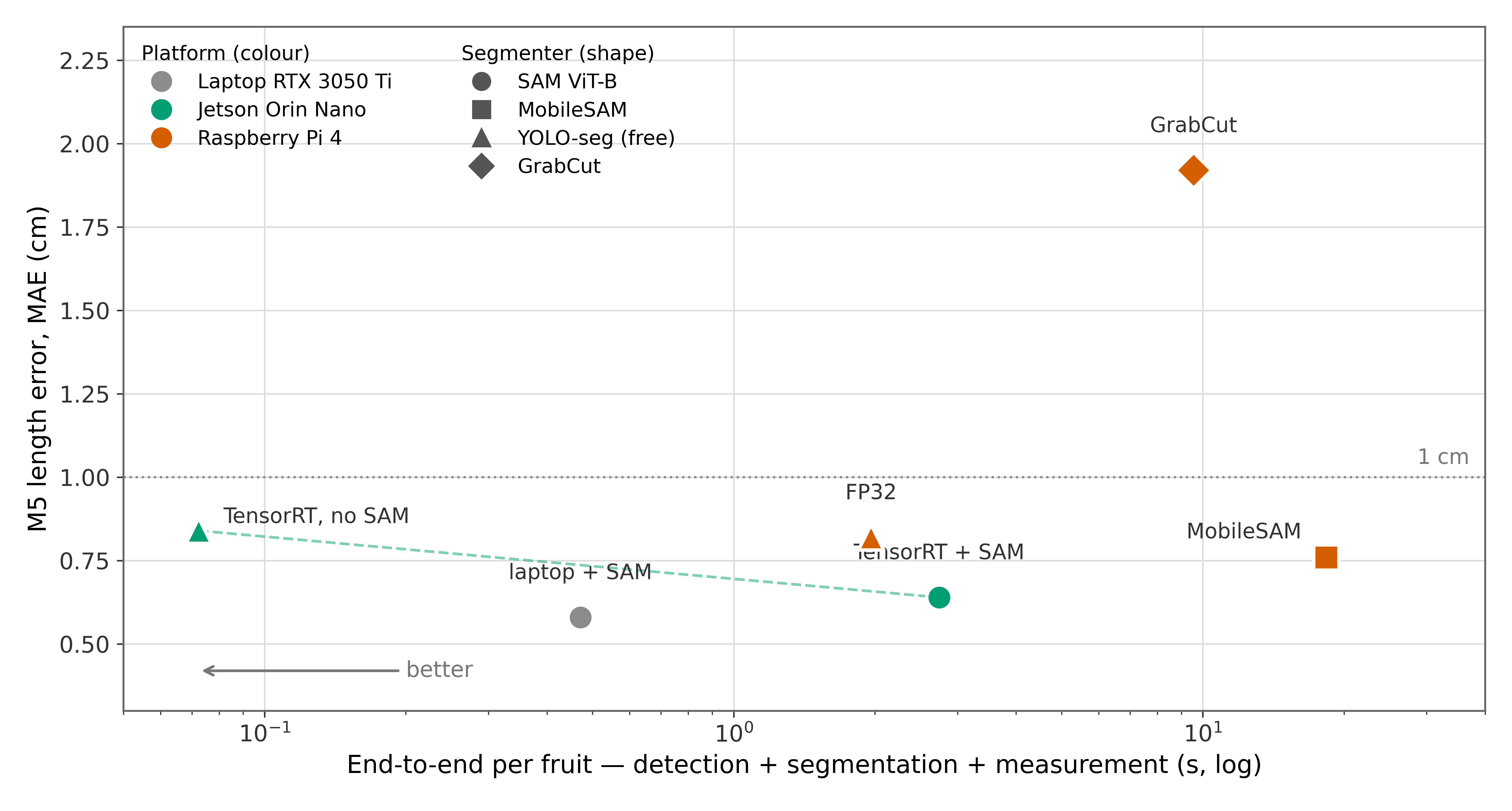}
\caption{The accuracy–throughput envelope. Each point is one platform and segmentation backend; colour denotes platform and shape denotes backend. The frontier is set by the mask source, and platform choice moves a configuration along the throughput axis without changing the accuracy the backend delivers.}
\label{fig:fig8}
\end{figure}

The practical outcome is an envelope rather than a single recommended configuration. \Cref{tab:pareto_edge} states it explicitly, with each configuration as a point in (accuracy, throughput, energy) space. Non-dominated configurations are those that no other configuration beats on both accuracy and throughput. Three are non-dominated: the Jetson with native masks for throughput, the Jetson with SAM for accuracy, and the Raspberry Pi with native masks as the low-cost option (\Cref{fig:fig8}).

\begin{table}[!ht]
\centering
\caption{The deployment envelope: accuracy--throughput--energy trade-off across configurations.}
\label{tab:pareto_edge}
\footnotesize
\setlength{\tabcolsep}{3pt}
\begin{threeparttable}
\begin{tabularx}{\linewidth}{@{}>{\raggedright\arraybackslash}X cccc@{}}
\toprule
\textbf{Configuration} &
\textbf{\shortstack{M5 MAE\\(cm)}} &
\textbf{\shortstack{Per fruit\\(s)}} &
\textbf{J/fruit} &
\textbf{\shortstack{Non-\\dominated}} \\
\midrule
Laptop, SAM ViT-B (reference)               & 0.58 & 0.47 & ---           & ref. \\
Jetson TensorRT FP16, native masks          & 0.84 & 0.07 & $\sim$0.7     & yes \\
RPi~4 INT8 head-FP32, native masks          & 1.12 & 1.44 & 11.3          & no  \\
RPi~4 FP32, native masks                    & 0.82 & 1.96 & 15.1          & yes \\
Jetson TensorRT FP16, SAM                   & 0.64 & 2.74 & 42.2          & yes \\
RPi~4 INT8, MobileSAM$^{\dagger}$           & 0.76 & 15.8 & $\sim$87.8    & no  \\
RPi~4 FP32, MobileSAM$^{\dagger}$           & 0.76 & 18.3 & 105.5         & no  \\
\bottomrule
\end{tabularx}
\begin{tablenotes}
\footnotesize
\item Energy is measured except where marked with a tilde ($\sim$); those entries are derived from per-fruit latency and the platform's mean measured draw (10.14\,W for the Jetson at 15\,W; 5.57\,W for the Raspberry Pi). Dominance is computed on accuracy and throughput only; a dominated configuration may still be the right choice on hardware cost, which this analysis does not model.
\end{tablenotes}
\end{threeparttable}
\end{table}

Three rules follow from the ablations, and they compose:
\begin{enumerate}
    \item Choose the segmentation backend for the accuracy target, on whatever hardware is convenient. For prompted backends, the accuracy transfers across platforms.
    \item Choose the platform for the throughput and energy target. The platform does not change the accuracy the backend delivers.
    \item Quantize the detector only where it is the bottleneck on this pipeline, the native-mask configurations and exclude the detection head when doing so.
\end{enumerate}

\subsection{The Structural INT8 Failure}

The failure is a property of the output layout, not of this model, this dataset or this toolchain. The precondition is a single tensor carrying both coordinate-scale and probability-scale quantities, quantized per tensor. That layout is characteristic of the NMS-free designs now displacing NMS-dependent heads~\citep{carion2020, zhao2024, wang2024yolov10, sapkota2025}. The practical recommendation is narrow: when quantizing a detector whose head is NMS-free, exclude the head and quantize the backbone. The head was 12.6\% of convolution weights here, so most of the benefit survives.

\subsection{Separability of Backend and Platform Selection}

That measurement accuracy is fixed by the segmentation backend and reproduces across unrelated silicon to within 0.01~cm for prompted backends, while the cost of that backend varies $214\times$ across the same platforms, is the most transferable finding here. A team can evaluate candidate segmentation backends on a workstation, where an experiment takes minutes rather than hours, and treat the resulting accuracy as what the field system will deliver.

Three boundary conditions apply. First, the property holds for prompted backends, not for the detector's own mask head. Second, the backends compared are deterministic given the same prompt. Third, the property holds above a mask-quality threshold: GrabCut breaks the separation by failing outright on a substantial fraction of frames.

\subsection{Limitations}

\textbf{Quantities not measured.} Mean average precision is not reported, as it requires independently annotated ground-truth boxes. Two entries in \cref{tab:pareto_edge} are derived from latency and mean draw rather than measured directly.

\textbf{Generalisation.} One unit of each board was tested; unit-to-unit variation is uncharacterised. Thermal characterisation was performed at laboratory ambient, and the extrapolation to greenhouse ambient is a single-condition inference.

\section{Conclusions and Future Work}

A curvature-aware RGB-D cucumber sizing pipeline was deployed on two classes of embedded hardware and characterised in detail. Both deployments succeed with accuracy preserved: M5 attains 0.58~cm (95\% CI 0.46--0.69) on the workstation baseline and 0.58~cm (95\% CI 0.46--0.69) on a Jetson Orin Nano, with a paired difference indistinguishable from zero. TensorRT FP16 compilation reduces Jetson detection latency $2.57\times$, yielding a 72~ms per-fruit operating point at 0.84~cm. Energy is 42.2~J per fruit on the Jetson with SAM at 15~W, and the passively cooled Raspberry Pi holds 1800~MHz without throttling through 30~min of continuous inference.

The ablations supply the configuration rules. Measurement accuracy is determined by the segmentation backend and reproduces across platforms to within 0.01~cm at matched precision for prompted backends, while segmentation cost is determined by the platform and varies 214-fold, making the two choices separable. Whole-graph INT8 quantization of an NMS-free detector fails structurally for a reason no calibration strategy can address; excluding the detection head restores function at $1.37\times$. A lower power mode is not necessarily a lower-energy one: the Jetson's 7~W setting costs 73\% more energy per fruit than its 15~W setting, and the lower-powered Raspberry Pi costs $2.5\times$ more per fruit than the Jetson at comparable accuracy.

Four directions follow. Scoring detection quality on-device with mean average precision (mAP) and mask intersection over union (IoU) against annotated ground truth would close the one accuracy question these results leave open. Field validation at greenhouse ambient, in the eventual enclosure, would test the thermal extrapolation of \cref{sec:thermal} against the condition that actually matters. Multi-fruit frames would exercise the segmentation stage differently from the single-fruit captures used here, making encoder reuse across prompts the dominant optimisation. Finally, dynamic power management that raises the power cap only while a measurement is in flight follows directly from the energy result.

\section*{Data and Code Availability}
Benchmark scripts, quantization configurations, the direct ONNX Runtime and TensorRT detectors, the energy and thermal harnesses, the statistical analysis and all result files are available from the corresponding author on request.

\bibliography{references}

\end{document}